%% file: main.tex
\input{header}

\title{Building LLM Agent Systems the Deep Learning Way: From Modular Design to Architecture Search}

\author{
  Tao Feng\thanks{Equal contribution.},
  Pengrui Han\footnotemark[1], Zhongjie Dai\footnotemark[1],
  Jiaxuan You\thanks{Corresponding author.}\\
  {\small University of Illinois Urbana-Champaign}\\
  {\addr{\{taofeng2, phan12, zhongjie, jiaxuan\}@illinois.edu}}
}

\begin{document}

\maketitle

\begin{abstract}
Large Language Models (LLMs) have revolutionized AI research and enabled exciting agent systems. To build a complex LLM agent system, most existing research relies on insights from other domains or heuristics to manually build the agent system. However, this approach often requires heavy hand-engineering and fails to fully optimize for the downstream task of interest. Inspired by the tremendous success of deep learning, we propose to construct LLM agent systems in a modular manner, similar to building a deep neural network. Our key insight is to make analogies between LLM building blocks, such as retrievals, memories, and prompting strategies, and the successful deep learning modules, such as MLPs, attention, and recurrent modules. We further design forward inference and feedback mechanisms for LLMs, where prompts in LLMs are considered as the weights in deep models, and the prompt optimization from feedback is analogous to the back-propagation algorithm. We additionally leverage a search algorithm to search for the best configuration of LLM agent systems, similar to the neural architecture search (NAS) in deep learning research. Comprehensive experimental results demonstrate that the proposed deep learning recipe for LLM agent systems is highly effective, in particular: (1) Organizing LLM modules into deep-learning-style architectures yields noticeable performance gain; (2) Automatic prompt optimization, equivalent to backpropagation, is efficient in incorporating feedback from the task of interest and achieves at least 5\% performance improvement; (3) NAS equivalent algorithm works well for further optimizing the LLM agent system architecture with 11\% performance gain compared with randomly designed architectures. Overall, our research demonstrates the exciting opportunity of transferring the success of deep learning to building LLM agent systems.
\end{abstract}
\input{sections/intro}
\input{sections/related}
\input{sections/methods}
\input{sections/expsetup}
\input{sections/exp}
\input{sections/discussion}

\bibliography{tmlr}
\bibliographystyle{tmlr}

\input{sections/appendix}

\end{document}

%% file: header.tex
\pdfoutput=1

\documentclass[10pt]{article}

\usepackage[preprint]{tmlr}
\usepackage{latexsym}
\usepackage{amsmath} 
\usepackage{amsthm}
 
\usepackage{mdframed}

\input{math_commands.tex}

\usepackage{hyperref}
\usepackage{url}

\usepackage{microtype}

\usepackage{inconsolata}

\usepackage{graphicx}

\usepackage{algorithm}
\usepackage{algorithmic}
\usepackage{url}
\usepackage{subfigure}
\usepackage{wrapfig}
\usepackage{booktabs}
\usepackage{multirow}
\usepackage{enumitem}
\usepackage{xspace}
\usepackage{listings}
\usepackage{tabu}
\usepackage{tabularx}
\usepackage{longtable}
\usepackage{microtype}
\newcommand{\para}[1]{{\vspace{3pt} \bf \noindent #1 \hspace{0pt}}}

\usepackage{rotating}
\usepackage{tcolorbox}
\usepackage{colortbl}

\usepackage{makecell}
\usepackage{bbding}
\usepackage{wrapfig}
\definecolor{royalblue(web)}{rgb}{0.25, 0.41, 0.88}
\definecolor{blue-violet}{rgb}{0.54, 0.17, 0.89}
\definecolor{brightmaroon}{rgb}{0.76, 0.13, 0.28}
\definecolor{darkmagenta}{rgb}{0.55, 0.0, 0.55}
\definecolor{bleudefrance}{rgb}{0.19, 0.55, 0.91}
\definecolor{palatinateblue}{rgb}{0.15, 0.23, 0.89}
\definecolor{royalblue(web)}{rgb}{0.25, 0.41, 0.88}
\definecolor{whitesmoke}{rgb}{0.96, 0.96, 0.96}
\definecolor{thulianpink}{rgb}{0.87, 0.44, 0.63}
\definecolor{amber(sae/ece)}{rgb}{1.0, 0.49, 0.0}
\definecolor{darkblue}{rgb}{0.0, 0.0, 0.55}
\definecolor{alizarin}{rgb}{0.82, 0.1, 0.26}
\definecolor{asparagus}{rgb}{0.53, 0.66, 0.42}
\definecolor{darkspringgreen}{rgb}{0.09, 0.45, 0.27}
\definecolor{columbiablue}{rgb}{0.61, 0.87, 1.0}
\definecolor{wildblueyonder}{rgb}{0.64, 0.68, 0.82}
\definecolor{trolleygrey}{rgb}{0.5, 0.5, 0.5}
\definecolor{paleaqua}{rgb}{0.74, 0.83, 0.9}
\definecolor{bubblegum}{rgb}{0.99, 0.76, 0.8}
\definecolor{coralred}{rgb}{1.0, 0.25, 0.25}
\definecolor{green(ryb)}{rgb}{0.4, 0.69, 0.2}
\definecolor{flame}{rgb}{0.89, 0.35, 0.13}
\definecolor{bittersweet}{rgb}{1.0, 0.44, 0.37}
\definecolor{darksalmon}{rgb}{0.91, 0.59, 0.48}
\definecolor{emerald}{rgb}{0.31, 0.78, 0.47}
\definecolor{green(pigment)}{rgb}{0.0, 0.65, 0.31}

\definecolor{codegreen}{rgb}{0,0.6,0}
\definecolor{codegray}{rgb}{0.5,0.5,0.5}
\definecolor{codepurple}{rgb}{0.58,0,0.82}
\definecolor{backcolour}{rgb}{0.96,0.96,0.94}

\definecolor{custompurple}{HTML}{800080}

\def\month{MM}  
\def\year{YYYY} 
\def\openreview{\url{https://openreview.net/forum?id=XXXX}} 

%% file: math_commands.tex
\usepackage{amsmath,amsfonts,bm}

\def\eqref#1{equation~\ref{#1}}

\def\1{\bm{1}}

\DeclareMathAlphabet{\mathsfit}{\encodingdefault}{\sfdefault}{m}{sl}
\SetMathAlphabet{\mathsfit}{bold}{\encodingdefault}{\sfdefault}{bx}{n}



%% file: sections/intro.tex
\section{Introduction}

Large Language Models (LLMs) have demonstrated remarkable potential in achieving Artificial General Intelligence (AGI) due to their impressive planning and reasoning abilities \citep{wu2023smartplay,ge2023llm}, which has sparked an upsurge in studies investigating sophisticated LLM agent systems  \citep{zhao2023expel,wang2023survey,peng2023check}. LLM agent systems usually involve various LLM building blocks  \citep{packer2023memgpt,yao2023react,topsakal2023creating,pandya2023automating,jeong2023generative}, such as retrieval \citep{robertson2009probabilistic,contriever,dragon}, memory modules \citep{zhao2023expel,wang2023survey}, and prompting strategies \citep{yao2024tree}. We argue that the success of LLM agent systems relies on the integration of multiple LLM modules \citep{zhao2023expel,talebirad2023multi,wu2023autogen}, which still receives limited attention from the community. Therefore, our paper aims to raise attention to the pressing research question: \textit{how to effectively and automatically construct LLM agent systems based on basic building blocks}.

Constructing effective LLM agent systems often involves leveraging insights from various disciplines, such as neuroscience \citep{yao2022react,shinn2023reflexion,kwon2023efficient} and computer architecture \mbox{\citep{packer2023memgpt,talebirad2023multi}}. For example, MEMGPT \citep{packer2023memgpt} draws inspiration from hierarchical memory systems in traditional operating systems to construct an LLM agent system with several manually designed LLM building blocks. However, applying such interdisciplinary insights to LLM agent system construction has the following shortcomings: (1) The approach is often domain-specific and not generic, requiring heavy manual designs; (2) Designing the LLM agent system necessitates extensive trial and error, making the process both costly and inflexible.

The tremendous success of Deep Learning (DL) has brought us a new perspective on solving the above problems. In the area of DL, researchers have proposed various basic modules such as Recurrent Neural Network (RNN) \citep{li2019deep,selvin2017stock,jordan1997serial} and Attention Network \citep{wang2017residual,fu2019dual,hou2019cross} to process various tasks. Constructing the basic modules into complicated architectures like Transformer \citep{vaswani2017attention,han2021transformer} and further leveraging the forward and back propagation \citep{johansson1991backpropagation,buscema1998back} to train such architectures have further contributed to the success of DL in tackling complex real-world applications. 

Inspired by the success of DL, we propose to \textit{design LLM agent systems with a Deep Learning recipe}. Our approach involves \textit{defining building blocks of LLMs analogous to the DL modules, constructing LLM agent systems similar to defining NN architectures, and developing the prompt optimization algorithm of LLMs similar to the backpropagation algorithm.}  Specifically, we explore LLM modules including: (1) vanilla LLM as a single layer MultiLayer Perceptron (MLP) \citep{popescu2009multilayer}; (2) Memory processer \citep{zhang2023memory,packer2023memgpt} as an RNN; (3) Retrieval \citep{robertson2009probabilistic,contriever,dragon} as an attention network \citep{wang2017residual,fu2019dual,hou2019cross}; (4) prompting strategies \citep{besta2023graph,yao2024tree} as Graph Neural Networks (GNNs) \citep{zhang2021magnet,shi2019skeleton,kenning2022directed}. Based on the defined LLM modules, we could construct more complex LLM agent systems by referencing successful DL architectures, notably, Transformer \citep{Vaswani+2017}. Furthermore, we design a meta prompt module similar to the training parameters in DL to guide the forward inference, and a feedback-propagation module that incorporates the external feedback and iteratively updates the meta prompt similar to the backpropagation algorithm. We additionally leverage a search algorithm to search for the best configuration of LLM agent systems, similar to the neural architecture search (NAS) \citep{liu2018progressive} in DL research.
In summary, our main contributions are as follows:
\begin{itemize}
\item  The first work to investigate constructing LLM agent systems from the DL perspective and recipe, extending the science and engineering of LLM agent systems.
\vspace{-0.5em}
\item Proposing DL-style building blocks for LLMs and constructing a ``Transformer'' architecture for LLM agent systems. Observing at least 5\% performance gain by ``backpropagating'' through the LLM agent system via a novel feedback-propagation algorithm.
\vspace{-0.5em}
\item  An automatic search algorithm for the best configuration of LLM agent systems, similar to NAS in DL, brings an 11\% performance gain compared to prior practice.  
\end{itemize}

%% file: sections/related.tex
\section{Related Works}
In this section, we first revisit the basic DL modules and how they are used to build complex NN architectures. We then review the current approaches in building LLM agent systems and discuss how we can bring insights from the success of DL applications to LLM agent systems.

\para{Modules of Deep Learning.} 
In the area of DL, researchers have designed various basic modules to tackle different tasks.  In the early period of DL design, researchers proposed MLP  \citep{popescu2009multilayer} to solve tasks such as classification, prediction, and regression. However, this basic architecture is less effective for complex input data. Consequently, researchers have developed specialized modules with various functionalities. RNNs \citep{li2019deep,selvin2017stock,jordan1997serial} are tailored for sequential data, maintaining a hidden state to capture information across time steps. CNNs \citep{kiranyaz2015real,kiranyaz20191} excel at capturing local patterns within images. Additionally, Attention mechanism \citep{wang2017residual,fu2019dual,hou2019cross} further refines model focus by emphasizing the most salient segments within the data. By combining and arranging these specialized modules, researchers have developed numerous powerful architectures. For instance, the Transformer \citep{vaswani2017attention,han2021transformer}, one of the most popular architectures today, relies on the attention mechanism and simple feed-forward layers. It has demonstrated surprising results in generative AI. Additionally, some researchers have integrated the attention mechanism into RNNs, yielding very promising outcomes \citep{wang2016survey,merity2019single}. Researchers have further developed automated architecture engineering, such as Neural Architecture Search (NAS) \citep{liu2018progressive,zoph2016neural,pham2018efficient}, aiming to find the optimized design of our machine learning model.

\para{Construction of LLM Agent Systems.} Much of the current research on LLM agent systems relies on the insights and observations from other domains \citep{yao2022react,shinn2023reflexion,kwon2023efficient,chen2023agentverse}.  However, these designs are often problem-specific and do not offer a systematic approach to defining a good LLM agent system architecture. Recent efforts, like Langchain \footnote{https://www.langchain.com} and LlamaIndex \footnote{https://www.llamaindex.ai}, aim to offer standardized building blocks and modules for LLM agent systems. However, they still lack a systematic approach to designing different LLM agent systems for various tasks, providing limited guidance on their integration for sophisticated and meaningful architectural compositions. In response, we propose the development of standardized modules for LLM agent system design, inspired by the proven methodologies within the DL domain. In this work, we show that these basic building blocks can be synthesized into a coherent framework, offering a novel, modular approach to architectures in LLMs that builds a bridge between LLMs and DL.

%% file: sections/methods.tex
\section{The Deep Learning Recipe for LLM Agent Systems}

To explore the deep learning recipe for LLM agent systems, we first introduce the basic building blocks of LLMs in Sec \ref{basic_block}, which demonstrates the relationships between DL and LLM building blocks.  Inspired by the success of DL's training process, we then illustrate the forward inference and feedback propagation of LLMs in Sec \ref{forward}. Finally, we introduce the instantiations of: building a ``Transformer'' with LLM building blocks in Sec \ref{instantiation}.




\begin{figure*}[t]
    \centering
    \vspace{-3mm}
    \includegraphics[width=\linewidth]{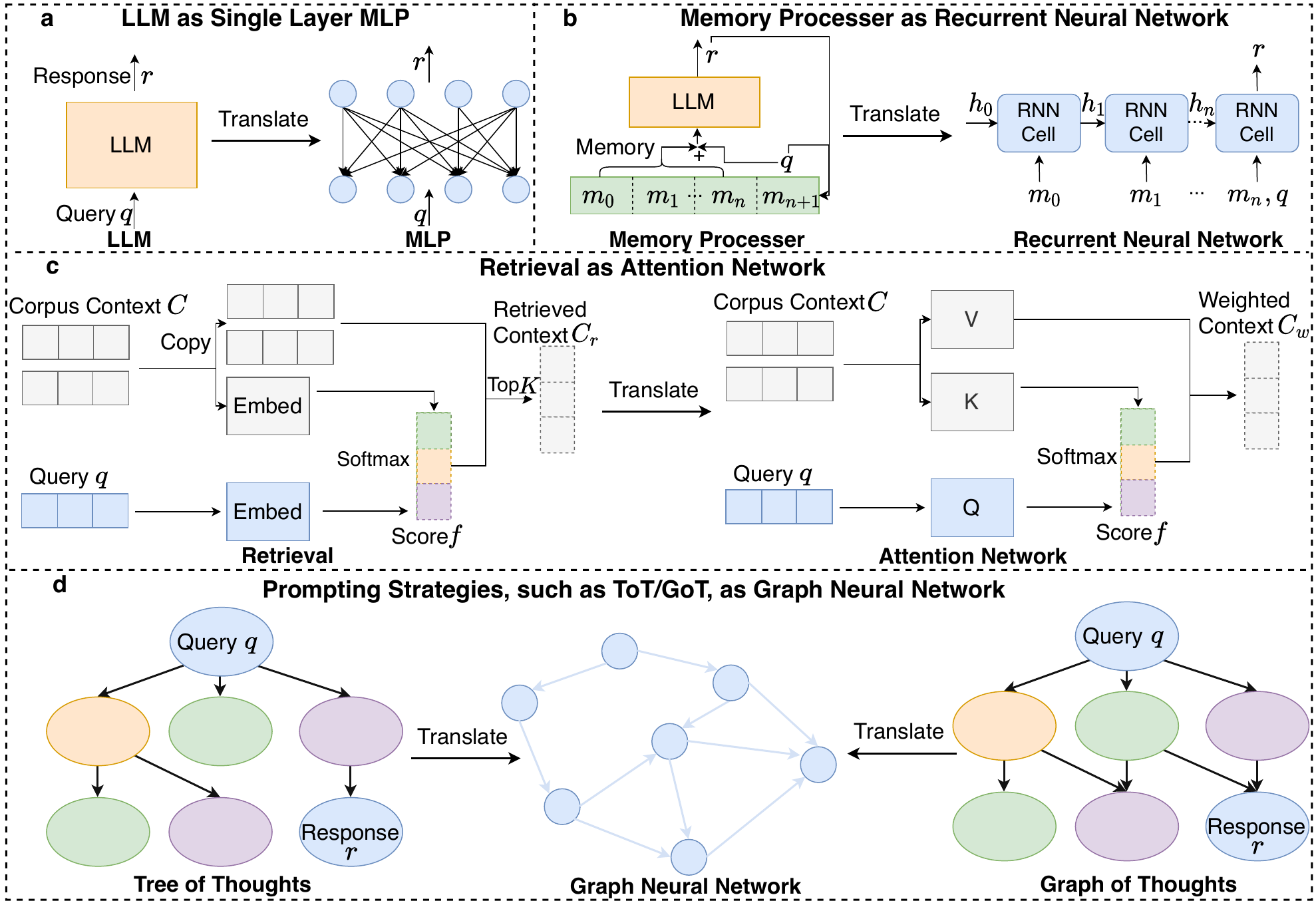}
    \caption{\textbf{Basic Building Blocks of LLM agent systems}. \textbf{(a)} A vanilla LLM can be translated into a single layer MLP, which receives a query $q$ and outputs a response $r$; \textbf{(b)} Memory processor can be regarded as RNN since they both utilize historical data for future planning;  \textbf{(c)} Retrieval is analogous to the attention network of DL because they both extract the importance information of the corpus context; \textbf{(d)} Prompting strategies, such as Tree/Graph of thoughts, are translated to  GNNs since they decompose query $q$ into interconnected thoughts, which are connected and related to each other.}
    \vspace{-3mm}
\label{fig:2.1}
\end{figure*}

\subsection{Deep-Learning-Style Building Blocks of LLM Agent Systems} \label{basic_block}

To introduce the basic building blocks of LLMs and the relationship between DL and LLMs, we ``translate'' the modules of LLMs into modules of DL as shown in Figure \ref{fig:2.1}.

\para{Vanilla LLM as Single Layer MLP.} As a basic component when constructing an LLM agent system, an LLM receives a query $q$ and outputs a response $r$. It can be translated as a single layer MLP, as shown in Figure \ref{fig:2.1}(a), which also plays a fundamental role in DL \citep{popescu2009multilayer}.

\para{Memory Processor as Recurrent Neural Network.} During the interaction process of the LLM, it will store the knowledge obtained by the interaction as memories $\{m_0,m_1,...m_n,q\}$ and utilize these memories for future responses \citep{shinn2023reflexion,kwon2023efficient} as shown in Figure \ref{fig:2.1}(b).  Similarly, Recurrent Neural Network (RNN) excels at processing sequential data due to its inherent structure that allows for the storage and utilization of previous information \citep{li2019deep,selvin2017stock,jordan1997serial}. It achieves this by maintaining hidden states $h_t$ that are updated at each time step $t$, effectively encoding the history of the sequence $\{m_0,m_1,...m_n,q\}$ up to that point. Therefore, we regard the memory processor as RNN.

\para{Retrieval as Attention Network.} When LLMs interact with factual knowledge stored in the corpus context $C$, they utilize retrieval \citep{robertson2009probabilistic,contriever,dragon} to efficiently retrieve accurate context for future responses. As shown in the left part of Figure \ref{fig:2.1}(c), to retrieve the accurate context, retrieval exploits the embedding similarity between query $q$ and corpus context $C$ and obtains the retrieved context $C_r$ of the top $K$ similarity ranking. This process is similar to the attention network shown in the right part of Figure \ref{fig:2.1}(c). They both utilize the embedding similarity between query $q$ and corpus context $C$ to obtain an importance score $f$, which determines the importance weights of each context. From this perspective, we translate the retrieval as an attention network. To be specific, single-query retrieval can be interpreted as single-head attention and multi-query retrieval can be regarded as multi-head attention \citep{wang2017residual,fu2019dual,hou2019cross}.

\para{Prompting Strategies as Graph Neural Network.} As shown in Figure \ref{fig:2.1}(d), common prompting strategies, such as Tree of Thoughts (ToT) \citep{yao2024tree} and Graph of Thoughts (GoT) \citep{besta2023graph} both decompose query $q$ into interconnected thoughts, which are connected and related to each other. These query and thought nodes and their relationships can be abstracted into nodes and edges of  Graph Neural Networks, respectively. Since a tree is a special case of a graph, we treat both ToT and GoT as instances of the GNN analogy in the subsequent introduction, denoting the corresponding architectures as Tree and GNN, respectively.

\begin{figure*}[t]
    \centering
    \vspace{-3mm}
    \includegraphics[width=\linewidth]{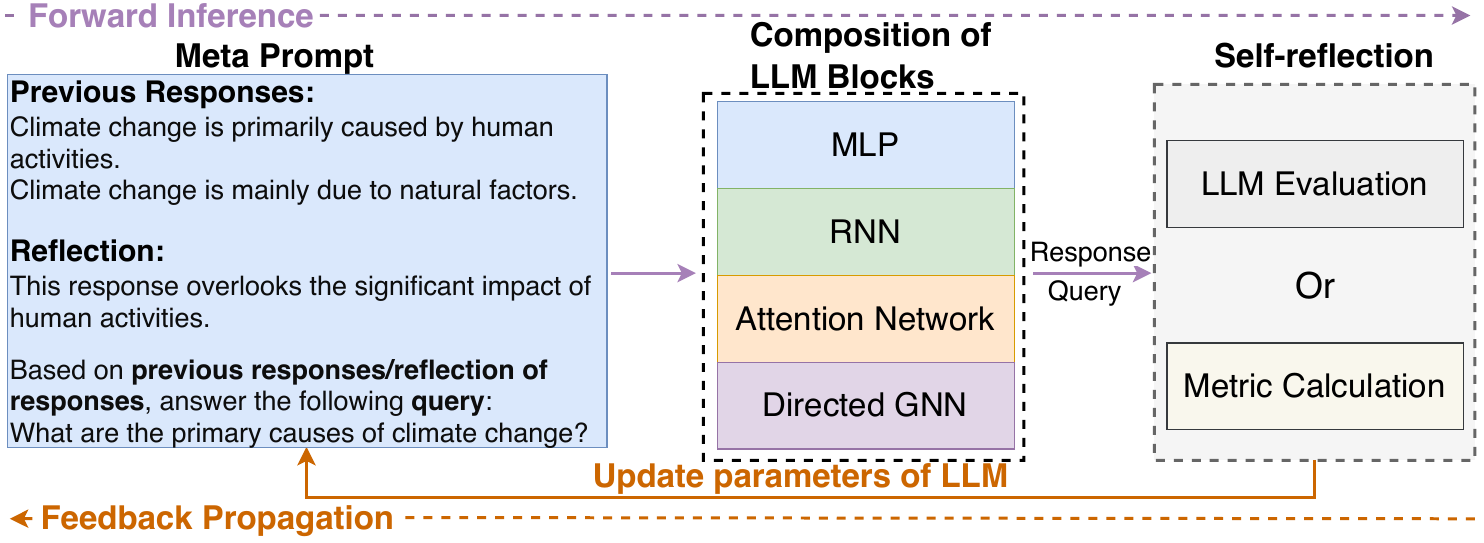}
    \caption{\textbf{Forward and Back Propagation Algorithms for Large Language Model}. In this process, meta prompt is regarded as trainable parameters in DL. For the forward inference, the input query goes through the composition of LLM building blocks to obtain the response, which is guided through the meta prompt. For the feedback propagation, the feedback from the self-reflection updates the parameters (meta prompt) of LLM.}
\label{fig:3.2}
\vspace{-3mm}
\end{figure*}

\subsection{Forward Inference and Feedback Propagation for LLM Agent Systems} \label{forward}
The success of DL not only lies in the design of fundamental modules discussed in Sec \ref{basic_block}, but also in the training process containing forward and back propagation \citep{buscema1998back} that brings external signals and feedback to the DL system. Inspired by this, we design the forward inference and feedback propagation of LLMs as shown in Figure \ref{fig:3.2}.


\para{Parameters of the LLM agent system.} The parameters can be divided into two categories: training parameters and hyperparameters. Here, the meta prompt in Figure \ref{fig:3.2} is equivalent to training parameters in deep learning, since it guides forward inference and can be updated through feedback propagation. Moreover, hyperparameters in LLM agent systems include max tokens of each LLM (size of MLP); number of retrievals (head number of attention network); retrieve number of each retrieval (mask); node number or depth of tree/graph of thoughts (node number or layer number of GNN); etc. These are predefined before using the LLM agent systems.

\para{Forward Inference.}
In the forward inference process of DL, the input data passes through each block of the neural network to obtain the final output \citep{buscema1998back,prabhushankar2022introspective}. Similar to this, in the forward inference process of LLM, the input query goes through the composition of LLM building blocks to obtain the response as shown in Figure \ref{fig:3.2}. Here we set up a meta prompt to guide the forward inference of LLM, which contains the input query, previous responses, and reflection returned by the self-reflection module.

\para{Feedback Propagation.} 
Many researchers attribute the success of DL to the feedback propagation process of DL \citep{buscema1998back,luft2014learning}, which updates the parameters of DL through feedback of losses between outputs and ground truths. Inspired by this, as illustrated in Figure  \ref{fig:3.2}, we regard the meta prompt as parameters of LLM and design a self-reflection module as the loss function to evaluate the effect of response on input query and update the parameters of LLM. Specifically, the self-reflection contains LLM Evaluation and Metric Calculation. For some queries whose response quality has clear metrics, we will choose Metric Calculation. Otherwise, we utilize LLM Evaluation to make LLM judge the quality of the response. The generated reflections and responses will be used to update the meta prompt to guide the LLM response in the next round.

\begin{figure*}[t]
    \centering
    \vspace{-3mm}
    \includegraphics[width=0.6\linewidth]{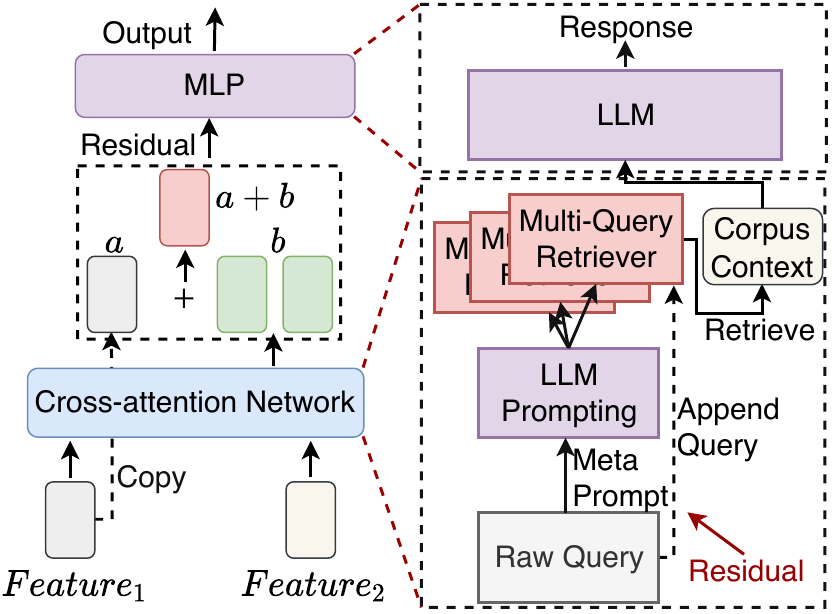}
    \caption{\textbf{Build a Cross-attention Transformer with LLM Building Blocks}. The multi-query retrieval is equivalent to the cross-attention network, and the LLM is translated to MLP in DL. Additionally, the process of appending the raw query to the generated multiple queries is equivalent to the residual.}
\label{fig:3.3}
\vspace{-3mm}
\end{figure*}

\subsection{Case study: Building  a ``Transformer'' with LLM Building Blocks} \label{instantiation}

In this section, we introduce how to build a (cross-attention) Transformer with LLM building blocks. As illustrated in Figure \ref{fig:3.3}, we show the correspondence between the main modules of the cross-attention transformer and LLM building blocks to illustrate how to utilize LLM building blocks to build a transformer. For the cross-attention transformer, the two different features of data will first go through a cross-attention network, and the output will be obtained through an MLP layer. To avoid the vanishing gradient problem or gradient explosion, researchers will also add the residual module to improve the training performance of the transformer \citep{vaswani2017attention,han2021transformer}. The success of the transformer in integrating and processing information from different features has inspired the design of LLM. As shown in the right part of Figure \ref{fig:3.3}, we analogize the raw query and the corpus context into two different features of data, like $Feature_1$ and $Feature_2$. On this basis, we first design a meta prompt containing raw queries to guide the LLM to generate multiple queries. Then we propose a multi-query retrieval based on multiple queries to retrieve relevant context from corpus context, which is equivalent to the multi-head cross-attention network. Additionally, we add the raw query to the generated multiple queries to ensure the effectiveness of our LLM, which is similar to the design of the residual module. Finally, an LLM will output the response like an MLP based on the retrieved context and input query. We summarize the DL recipe for LLM agent systems as a user instruction manual in Table \ref{tab:Instruction Manual} and please refer to  Appendix \ref{Manual} for details.

%% file: sections/expsetup.tex

\section{Experimental Setup}


\subsection{Datasets and Tasks}
We evaluate LLM agent systems on two major categories of tasks. They differ in whether the LLMs interact with external knowledge \citep{sun2023head}, such as external factual knowledge. Table \ref{tab:dataset} in Appendix \ref{Manual} summarizes the datasets and the number of cases used for evaluation.

\paragraph{LLM without External Knowledge.} We evaluate non-retrieval-based methods using two datasets: (1)\textbf{Traveling Salesman Problem (TSP)} \citep{hahsler2007tsp},  where we create instances by randomly generating $n$ nodes with $x$ and $y$ coordinates in the $[-10,10]$ range. As a classic combinatorial optimization problem, performance is evaluated by the length of the best solution within the fixed exploration step. (2) \textbf{GSM8k}  \citep{cobbe2021training}, which contains high-quality, linguistically diverse grade-school math word problems necessitating multi-step reasoning, crafted by expert problem writers. Each problem includes a detailed reasoning process and the correct final answer, with performance evaluated based on final answer accuracy.

\paragraph{LLM with External Knowledge.} Our evaluation of retrieval-based frameworks utilizes two datasets: (1) \textbf{StrategyQA} \citep{geva2021did}, a benchmark for open-domain commonsense reasoning question-answering that highlights the importance of implicit reasoning. It involves retrieving pertinent evidence paragraphs from a corpus to correctly answer questions. We compare the retrieved paragraphs with the ground truth and obtain Precision, Recall, and F1 as evaluation metrics.
(2) \textbf{Related-multi} (will be released after review), designed to assess a language model's capability to effectively use extensive context from real-world academic papers for answering queries. Specifically, the task is to write a related work section based on the title and abstract of a target paper. LLM needs to use the title and abstract as the query to retrieve text chunks to complete this task. Text chunks depict the abstracts of several papers (each text chunk corresponds to the abstract of a paper), where some papers are cited in the related work section of the target paper, while others are randomly sampled from the same broader field. The retrieved text chunks are compared with the ground truth to obtain Precision, Recall, and F1 as evaluation metrics.

\begin{figure*}[t]
    \centering
    \vspace{-3mm}
    \includegraphics[width=0.9\linewidth]{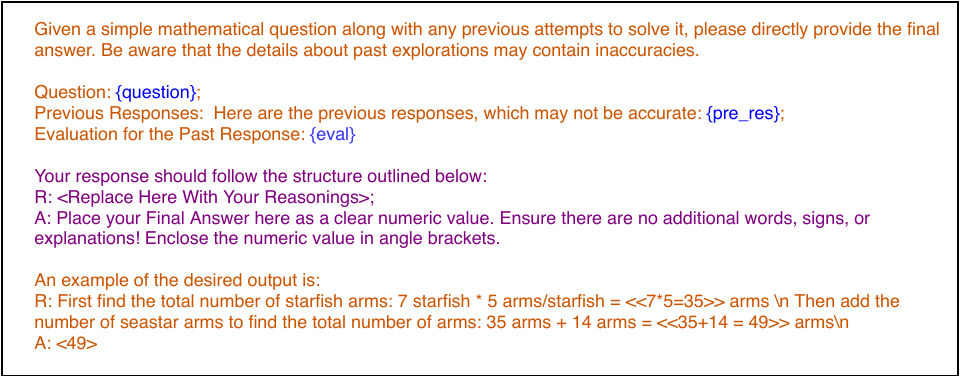}
    \caption{An example of the meta prompt used for RNN-MLP on the GSM8K dataset. The \textcolor{blue}{blue} text contains input questions and previous responses/evaluations. The \textcolor{custompurple}{purple} text describes the output format instructions.
}
\label{fig:4.2}
\vspace{-3mm}
\end{figure*}

\subsection{Models: Various Architectures for LLM Agent Systems}

We delve into the detailed descriptions of the architectures we explored, with a particular emphasis on their implementations and the methods employed for backward propagation across various modules. For the two LLM datasets without external knowledge, we design the following four basic architectures: (1) \textbf{MLP:} a single LLM; (2) \textbf{RNN-MLP:} previous responses are appended to the meta prompt to guide the generation of response; (3) \textbf{Tree-MLP:} utilizes ToT to generate multiple thoughts for response generation; (4) \textbf{GNN-MLP:} GoT is exploited to develop various thoughts for response generation.

We also develop four basic architectures for LLM agent systems with external knowledge. In order to interact with external knowledge, the four architectures are supplemented with the attention to retrieve relevant information based on the above architectures: (1) \textbf{Att-MLP:} a single retrieval to retrieve relevant information for LLM; (2) \textbf{RNN-Att-MLP:} previous responses are appended into the meta prompt to change the query for retrieval; (3) \textbf{Tree-Att-MLP:} utilize ToT to generate new query for retrieval; (4) \textbf{GNN-Att-MLP:} GoT is exploited to develop new query to retrieve. More details about the basic architectures of LLM agent systems are introduced in Appendix \ref{baseline}. Additionally, we have shown an example of the meta prompt used for RNN-MLP on the GSM8K dataset in Figure \ref{fig:4.2}.  Other meta prompts are also summarized in Appendix \ref{prompts}.

%% file: sections/exp.tex
\section{Results}

\begin{table*}[t]
    \begin{minipage}[b]{0.37\textwidth}
        \centering
        \caption{\textbf{Performance of Different LLM agent systems on TSP and GSM8k}.}
        \renewcommand{\arraystretch}{2.0} 
        \resizebox{\textwidth}{!}{%
        \begin{tabular}{c||c|c}
            \Xhline{1pt} 
            \rowcolor{paleaqua}
            \textbf{Type} & \multicolumn{2}{c}{\textbf{Without External Knowledge}}  \\
            \hline
            \rowcolor{paleaqua}
            \textbf{Dataset} & \multicolumn{1}{c|}{\textbf{TSP}} & \multicolumn{1}{c}{\textbf{GSM8k}} \\
            
            \rowcolor{paleaqua}
            Method  & Length & Accuracy  \\
            \hline \hline
            
            MLP & 79.2 & 53.3\%  \\
            \hline 
            \rowcolor{gray!20}
            RNN-MLP  & 76.6 & 60.0\%  \\

            Tree-MLP  & 73.2 &  66.7\% \\
    
            \rowcolor{gray!20}
            GNN-MLP  & 73.9 &  64.3\% \\
    
            \Xhline{1pt} 
        \end{tabular}
        }
        \label{tab:main result_1}
    \end{minipage}
    \hfill
    \begin{minipage}[b]{0.6\textwidth}
        \centering
        \caption{\textbf{Performance of Different LLM agent systems on StrategeyQA and Related-multi}. }
        \renewcommand{\arraystretch}{2.0} 
        \resizebox{\textwidth}{!}{%
        \begin{tabular}{c||ccc|ccc}
            \Xhline{1pt} 
            \rowcolor{paleaqua}
            \textbf{Type} & \multicolumn{6}{c}{\textbf{With External Knowledge}}  \\
            \hline
            \rowcolor{paleaqua}
            \textbf{Dataset} & \multicolumn{3}{c|}{\textbf{StrategeyQA}} &  \multicolumn{3}{c}{\textbf{Related-multi}} \\
            \rowcolor{paleaqua}
            Method & Precision & Recall & F1& Precision & Recall & F1   \\
            \hline \hline
            
            Att-MLP  & 0.45 & 0.40 & 0.42 & 0.31  & 0.17 & 0.22   \\
    
            \rowcolor{gray!20}
            RNN-Att-MLP  & 0.72 & 0.62& 0.66 & 0.33 & 0.19  & 0.24    \\

            Tree-Att-MLP  & 0.53 & 0.47 & 0.50 &0.34  & 0.20 & 0.25  \\
    
            \rowcolor{gray!20}
            GNN-Att-MLP  & 0.58 &  0.51 & 0.54 & 0.29 & 0.16 & 0.21   \\
    
            \Xhline{1pt} 
        \end{tabular}
        }
        \label{tab:main result_2}
    \end{minipage}
\end{table*}

\subsection{Performance of Different Architectures of LLM Agent Systems}

\para{Different Tasks Call for Different LLM Architectures.}
Our results across two distinct types of datasets are illustrated in Tables \ref{tab:main result_1} and \ref{tab:main result_2}, demonstrating that the integration of RNN, Tree, and GNN generally enhances the MLP's performance across various tasks. Specifically, in reasoning (GSM8K) and optimization (TSP) problems that do not require external knowledge, GNN and Tree excel due to their strong logical capabilities. For tasks involving reasoning (StrategyQA) and long-context summary (Related-multi) that necessitate external knowledge, RNNs proved most effective, in line with our expectations. This effectiveness is attributed to their capability to process and retain long sequences of information. Consistent with findings in DL, combining RNNs with attention mechanisms has been shown to yield promising results.

\para{Feedback Propagation Effectively Optimizes LLM Agent System Performance without Human Intervention.} Through our experiments, we discover that our automatic prompt optimization technique, akin to backward propagation, consistently improves the performance of the LLM agent systems in most scenarios. This verifies that LLM agent systems benefit from integrating past experiences and self-evaluations. Figure \ref{fig:model_comparison} presents the results at different steps on the TSP datasets, illustrating that feedback propagation can bring at least 5\% performance gain for LLM agent systems. Moreover, we can also observe that GNN-MLP and Tree-MLP iterate faster than other architectures and obtain better results. This is because they explore more solutions at each step and improve the efficiency of optimization.

\begin{wrapfigure}{r}{5.5cm}
\vspace{-1.5em}
\includegraphics[width=5.2cm]{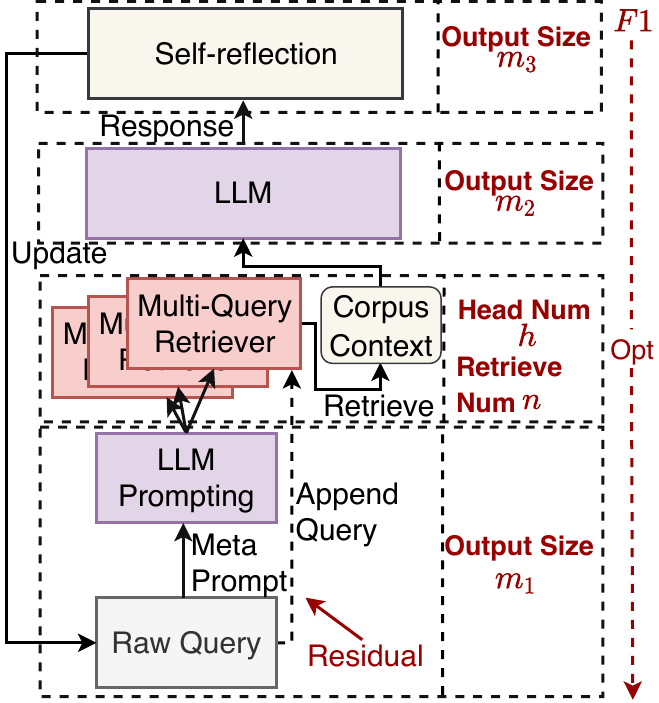}
\caption{\textbf{Optimization on transformer to maximize F1.}}
\label{fig:exp}
\vspace{-1.2em}
\end{wrapfigure}

\subsection{Case Study: Results of ``Transformer-style'' LLM Agent Systems}

\begin{figure}[t]
    \centering
    \begin{minipage}{0.47\linewidth}
\vspace{-3mm}
        \includegraphics[width=\linewidth]{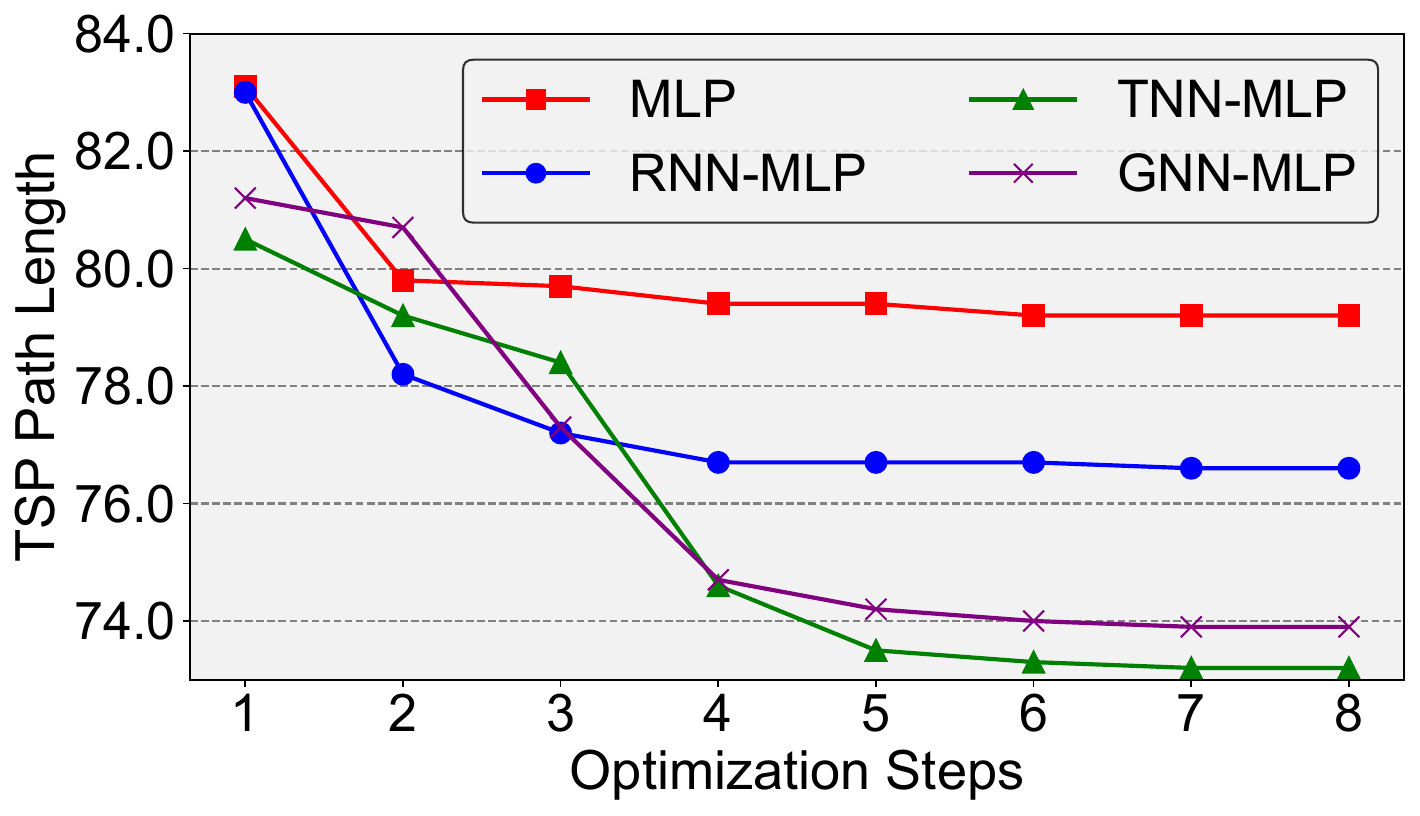}
        \caption{\textbf{The performance of LLM agent systems on the TSP dataset will get better as the feedback propagation progresses.} }
        \label{fig:model_comparison}
\vspace{-3mm}
    \end{minipage}
    \hfill 
    \begin{minipage}{0.49\linewidth}
\vspace{-3mm}
        \includegraphics[width=\linewidth]{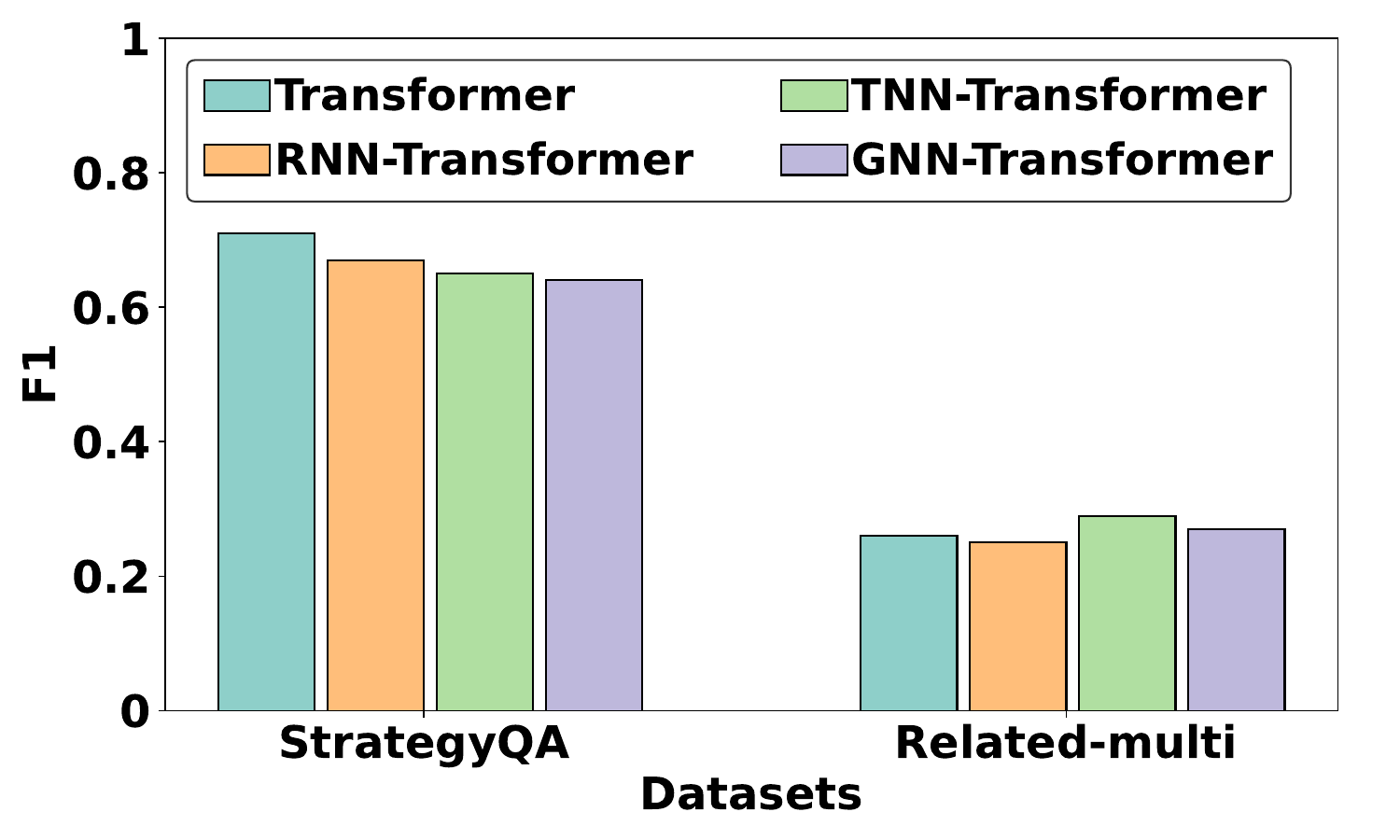}
        \caption{\textbf{Case Study: Using LLM Building Blocks to Build Variants of Transformer}.}
        \label{fig:case_study}
\vspace{-3mm}
    \end{minipage}
\end{figure}

We utilize variants of the transformer as an example to show the role of different LLM building blocks when building complex architectures. To be specific, we set up the following variants. (1) \textbf{Transformer}: It denotes transformer  illustrated in Sec \ref{instantiation}. (2) \textbf{RNN-Transformer}: We add previous responses to the meta prompt of the transformer to utilize its memory, which is commonly used in DL \citep{xia2019rthn,liu2019r}  when processing historical data. (3) \textbf{Tree-Transformer/GNN-Transformer}: ToT/GoT is added to transformer to guide multi-query generation, which is similar to GraphTransformer \citep{yun2019graph,hu2020heterogeneous,rampavsek2022recipe} in DL. We compare their performance on two datasets with the same hyperparameters and report all the results in Figure \ref{fig:case_study}. It can be observed that the Transformer performs best in StrategyQA. This is because StrategyQA's tasks are relatively simple, and Transformer is already relatively complex, therefore other variants do not perform well. We can also observe that Tree-Transformer and GNN-Transformer perform relatively better than other variants on Related-multi. It is because writing related work requires more thinking and divergence of thoughts \citep{randolph2019guide,torraco2005writing}.

\subsection{Automatic Architecture Search of LLM Agent Systems}
To automatically optimize the LLM agent system architecture like NAS, we propose an optimization framework based on Optuna \citep{akiba2019optuna}, which is an efficient and lightweight hyperparameter optimization software. To better illustrate our framework, we take the optimization of the transformer introduced in Sec \ref{instantiation} on Related-multi as an example. As shown in Figure \ref{fig:exp}, we optimize the hyperparameters of each module using Optuna. To be specific, we first conclude the hyperparameters and define their optimization range in Table \ref{optimization range} of Appendix \ref{opt}: Output size $m_1$ (max tokens) for LLM prompting, head num $h$ and retrieve num $n$ of each head for multi-retrieval, output size $m_2$ for output LLM, and output size $m_3$ for self-reflection. We utilize 30\% of the dataset (training set) for searching for an optimization architecture and 70\% for testing (testing set) and further define  F1 as the optimization goal. For each set of hyperparameters  $pr_{i}=(m_1,m_2,m_3,h,n)$, we calculate its F1 performance on training set. As illustrated in Figure \ref{fig:NAS} of Appendix \ref{opt}, we can obtain $\{pr_{i}, F1_{i}\}^k$  via  $k$ trials (120 trials in our experiment) and optimize the architecture by maximizing F1. In addition, as shown in Table \ref{table:opt_effect}, we select some points to illustrate that the LLM agent system becomes better as the number of trials increases. Through the experiment, we obtain the optimized hyperparameters    $pr=(m_1=192,m_2=192,m_3=320,h=2,n=4)$ with $F1=0.712$ on the testing set, which has 11\% better performance than randomly designed architectures (obtained by conducting experiments in which $m$ groups of hyperparameters were randomly selected and averaging the corresponding F1 scores).

\begin{wraptable}{r}{0.45\textwidth} 
\vspace{-2em} 
\caption{\textbf{As the number of LLM architecture searches increases, the LLM agent system is getting better.}}
\label{table:opt_effect}
\centering
\renewcommand{\arraystretch}{1.2} 
\setlength{\tabcolsep}{4pt} 
\begin{tabular}{p{2.5cm}cccc} 
\toprule
 & \textbf{0} & \textbf{7} & \textbf{43} & \textbf{78} \\ \midrule
F1 Score & 0.541 & 0.685 & 0.694 & 0.712 \\ \bottomrule
\end{tabular}
\vspace{-1em} 
\end{wraptable}

%% file: sections/discussion.tex
\section{Conclusion and Discussion}
\paragraph{Conclusion.} In conclusion, our work introduces a pioneering modular approach to the construction of LLM agent systems, establishing parallels with DL. We provide comprehensive methods and insights into viewing and assembling LLM building blocks in a manner analogous to DL architectures. Through extensive experimentation, we demonstrate that specific classic LLM modules are optimized for distinct tasks, similar to those in DL, and can be integrated in a manner reminiscent of DL configurations. Moreover, our feedback system enhances the performance of LLMs, paralleling the training mechanisms in DL. Finally, we propose an automatic optimization method for identifying the most effective LLM architectures, presenting a promising avenue for future research. 

\paragraph{Discussion.} This work represents a pioneering effort in the development of systematic and scientifically grounded LLM agent systems, drawing upon DL insights to inaugurate a new era of efficiency and standardization in the research and application of LLMs. By establishing fundamental LLM agent systems, our objective is to reconcile the existing diversity of paradigms within LLM research and, akin to DL, introduce classic and standardized modules that enhance scientific rigor, effectiveness, and innovation within the academic community. In the industrial context, the adoption of foundational and systematic architectures enables the creation of more efficient LLM building blocks, thus establishing a solid basis for the field of LLM agent systems. These advanced LLM building blocks possess the potential to streamline the deployment of LLM products, thereby enabling significant real-world impact and facilitating transformative changes across various sectors. Inspired by the progression of DL, this work introduces a suite of standardized modules within the LLM agent system, setting a foundation for future research to explore multi-LLM settings, integrate more established DL modules and architectures like CNN, and adopt self-exploration algorithms akin to NAS to revolutionize the optimization of LLM agent systems, potentially accelerating the discovery of innovative solutions.

%% file: sections/appendix.tex
\clearpage
\appendix

\section{User Instruction Manual} \label{Manual}
 
\begin{table}[h]
\begin{center}
	\caption{Deep Learning Recipe for LLM Agent Systems.}
	\resizebox{5.4in}{!}{
	\begin{tabu}{l|l}
        \hline
		\textbf{LLM agent systems} & \textbf{Deep Learning Architecture} \\ \hline
	   Vanilla LLM & Single layer MLP\\
		 Memory processor &  Recurrent neural network\\
        Retrieval &  Attention Network\\
    Prompting strategies (ToT \& GoT) &  Graph neural network\\
     Prompts &  Training parameters\\
       Max tokens of each LLM & Size of MLP\\
         Number of retrievals &  Head number of attention  network\\
        Node number or depth of tree/graph &  Node number or layer number of GNN\\
        Forward inference of LLMs & Forward inference  \\
        Feedback propagation & Back propagation   \\
        Self-reflection  & Loss function  \\
        Retrieve number of each retrieval & Mask \\\hline
	\end{tabu}}
	\label{tab:Instruction Manual}
\end{center}
\end{table}

\section{Prompts} \label{prompts}

\begin{figure*}[h]
    \centering
    \includegraphics[width=1\linewidth]{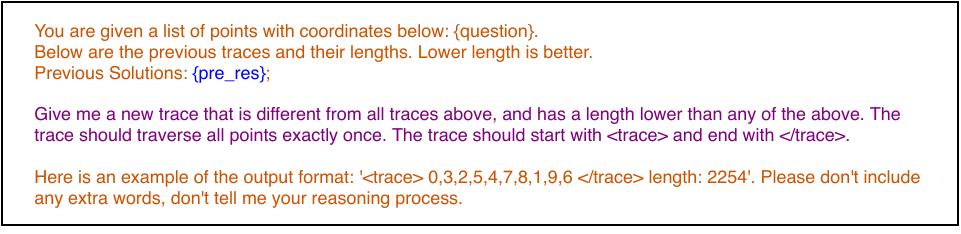}
    \caption{An example of the meta prompt used for RNN-MLP on the TSP dataset. The \textcolor{blue}{blue} text contains input questions and previous responses/evaluations. The \textcolor{custompurple}{purple} text describes the output format instructions.
}
\label{fig:9.1}
\end{figure*}

\begin{figure*}[h]
    \centering
    \includegraphics[width=1\linewidth]{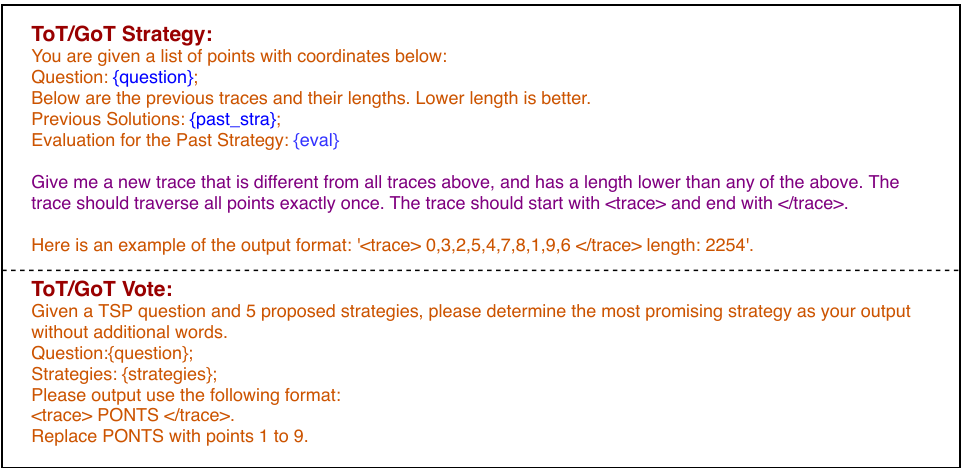}
    \caption{An example of the meta prompt used for Tree-MLP/GNN-MLP on the TSP dataset. The \textcolor{blue}{blue} text contains input questions and previous responses/evaluations. The \textcolor{custompurple}{purple} text describes the output format instructions.
}
\label{fig:9.2}
\end{figure*}

\begin{figure*}[ht]
    \centering
    \includegraphics[width=1\linewidth]{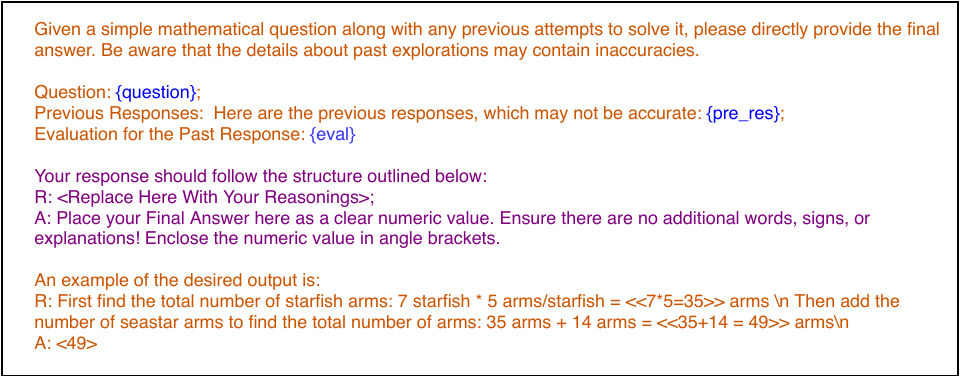}
    \caption{An example of the meta prompt used for RNN-MLP on GSM8K dataset. The \textcolor{blue}{blue} text contains input questions and previous responses/evaluations. The \textcolor{custompurple}{purple} text describes the output format instructions.
}
\label{fig:9.3}
\end{figure*}

\begin{figure*}[ht]
    \centering
    \includegraphics[width=1\linewidth]{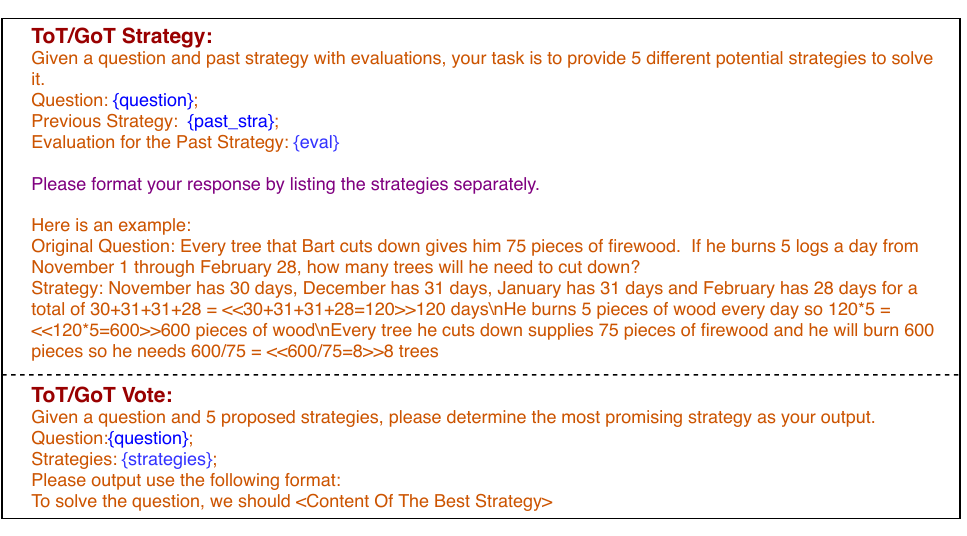}
    \caption{An example of the meta prompt used for Tree-MLP/GNN-MLP on GSM8K dataset. The \textcolor{blue}{blue} text contains input questions and previous responses/evaluations. The \textcolor{custompurple}{purple} text describes the output format instructions.
}
\label{fig:9.4}
\end{figure*}

\begin{figure*}[ht]
    \centering
    \includegraphics[width=1\linewidth]{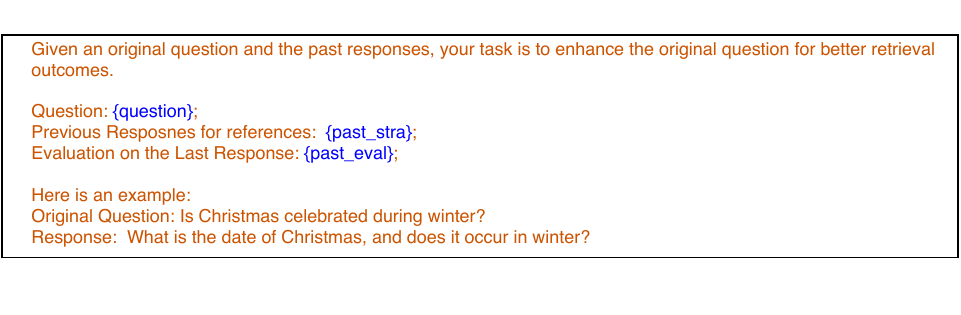}
    \caption{An example of the meta prompt used for RNN-Att-MLP on StrategeyQA dataset. The \textcolor{blue}{blue} text contains input questions and previous responses/evaluations. The \textcolor{custompurple}{purple} text describes the output format instructions.
}
\label{fig:9.5}
\end{figure*}

\begin{figure*}[h]
    \centering
    \includegraphics[width=1\linewidth]{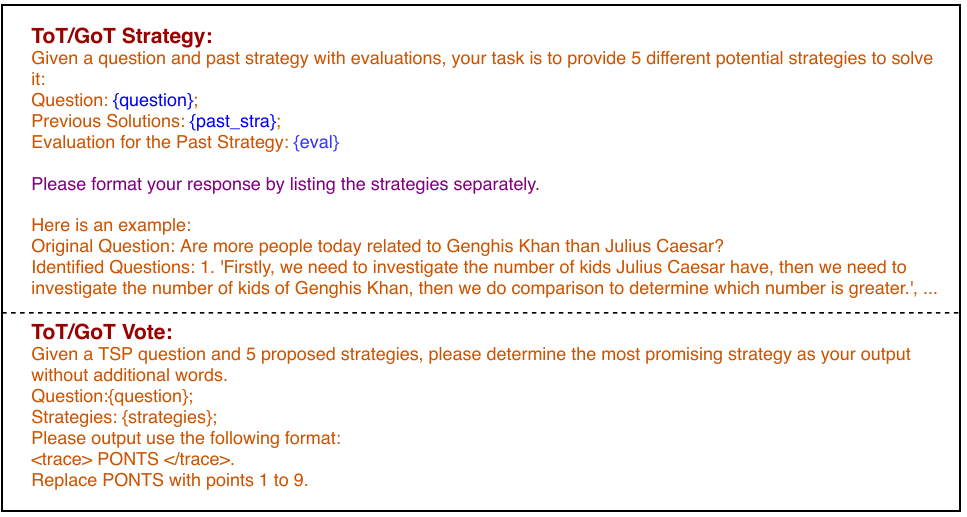}
    \caption{An example of the meta prompt used for Tree-Att-MLP/GNN-Att-MLP  on StrategeyQA dataset. The \textcolor{blue}{blue} text contains input questions and previous responses/evaluations. The \textcolor{custompurple}{purple} text describes the output format instructions.
}
\label{fig:9.6}
\end{figure*}

\begin{figure*}[ht]
    \centering
    \includegraphics[width=1\linewidth]{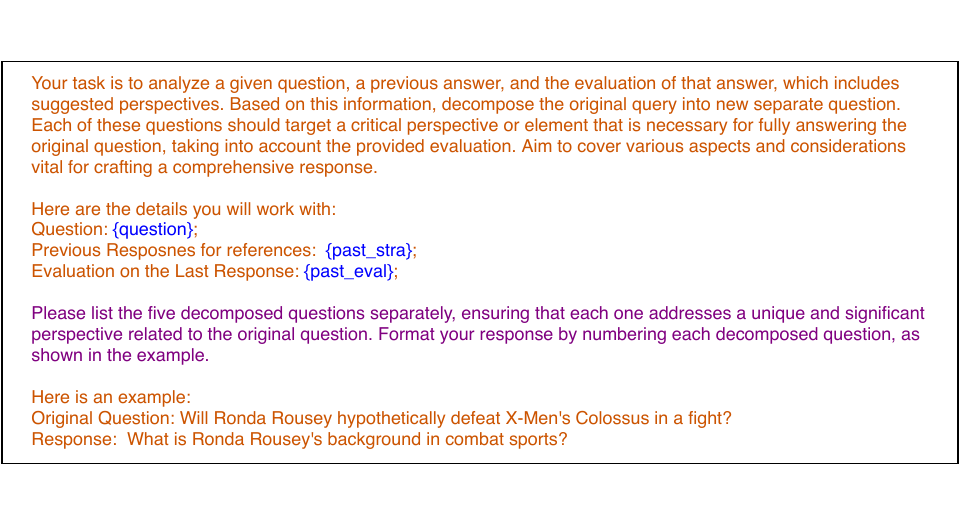}
    \caption{An example of the meta prompt used for RNN-Att-MLP on Related-multi dataset. The \textcolor{blue}{blue} text contains input questions and previous responses/evaluations. The \textcolor{custompurple}{purple} text describes the output format instructions.
}
\label{fig:9.7}
\end{figure*}

\begin{figure*}[ht]
    \centering
    \includegraphics[width=1\linewidth]{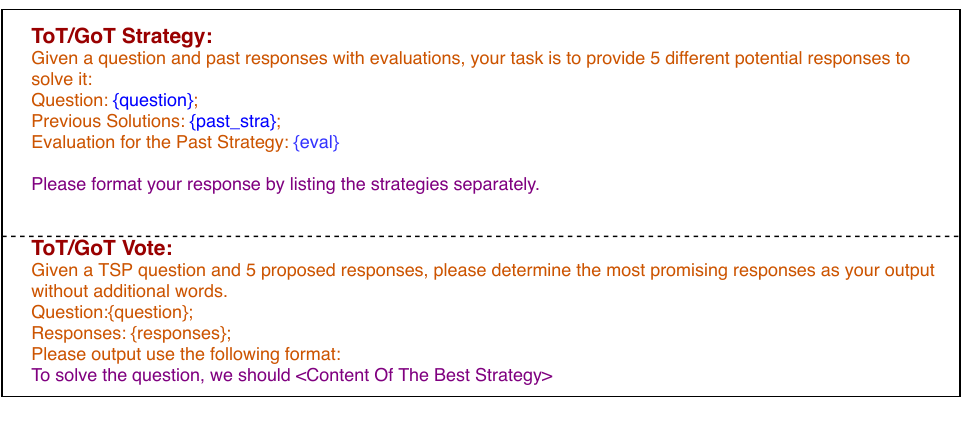}
    \caption{An example of the meta prompt used for Tree-Att-MLP/GNN-Att-MLP on Related-multi  dataset. The \textcolor{blue}{blue} text contains input questions and previous responses/evaluations. The \textcolor{custompurple}{purple} text describes the output format instructions.
}
\label{fig:9.8}
\end{figure*}




\clearpage

\section{Details of Architectures for LLM Agent Systems} \label{baseline}

\paragraph{MLP.} For the single-layer MLP setup, as previously defined, the LLM processes an input query $q$ and generates a corresponding output response $r$. Subsequently, through a process of backpropagation, the LLM undertakes a self-evaluation mechanism. The outcomes of this self-evaluation $e$ and response $r$ are then integrated into the original $q$, facilitating an updated $r$. 

\paragraph{RNN-MLP.} In this RNN+MLP framework, the model processes a query \(q\) and a sequence of past experiences \(m_0, m_1, \ldots, m_n\) with evaluations \(e_0,e_1,\ldots,e_n\). At each step, the latest three experiences \(m_{n-2}, m_{n-1}, m_n\) and evaluations \(e_{n-2}, e_{n-1}, e_n\) are integrated into the query \(q\). This enriched query is then utilized to generate the next response \(m_{n+1}\). After that, we evaluate it to get $e_{n+1}$.  \(m_{n+1}\) and $e_{n+1}$ are then fed back through backward propagation for subsequent iterations. Figure \ref{fig:4.2} shows the example meta prompt for how RNN+MLP works on GSM8K. 

\paragraph{Tree-MLP.} In this setting, we use zero-shot ToT. Specifically, a query \(q\) and a meta prompt \(q_m\) for decomposing \(q\) into interconnected thoughts are used to construct a comprehensive prompt \(p\). This prompt serves as input for an LLM, yielding a response \(r\). During backward propagation, a self-evaluation is performed, and the resulting evaluation \(e\) is utilized to refine \(q_m\) into an updated version \(q_m'\), leading to the formation of an updated prompt \(p'\). 

\paragraph{GNN-MLP.} This is similar to the Tree-MLP architecture, with the key difference that we allow for deeper graph-based thoughts.

\paragraph{Att-MLP.} In our single retrieval Att-MLP model, an initial input query $q$ and a meta query $q_m$ are used to generate an enhanced query $p$. Information $Cr$ is retrieved from context $C$ based on $p$, and both $q$ and $Cr$ are inputted into an LLM (MLP) to generate a response $r$. A feedback system evaluates this response, producing an evaluation $e$. Both $r$ and $e$ are then fed to $q_m$ which is then utilized to update $p$.

\paragraph{RNN-Att-MLP.} In single retrieval with RNN, the model processes an input query \(q\), a meta query \(q_m\), and a sequence of past experiences \(m_0, m_1, \ldots, m_n\) alongside their evaluations \(e_0, e_1, \ldots, e_n\). At each step, the latest three experiences \(m_{n-2}, m_{n-1}, m_n\) and their evaluations \(e_{n-2}, e_{n-1}, e_n\), combined with the input query \(q\), are integrated into the meta query \(q_m\) to generate an enhanced query \(p\). Information \(Cr\) is then retrieved from context \(C\) based on \(p\), and both \(q\) and \(Cr\) are fed into an LLM (MLP) to produce a response \(m_{n+1}\). This response is evaluated by a feedback system, generating an evaluation \(e_{n+1}\). Both \(m_{n+1}\) and \(e_{n+1}\) are stored for future use.

\paragraph{Tree-Att-MLP.} The Tree-Att-MLP configuration implements zero-shot ToT before the attention module. A query \(q\), a meta prompt \(q_m\), and a meta query \(q_n\) are provided. Initially, \(q\) is processed through \(q_m\) to decompose into interconnected thoughts \(T_1,\ldots,T_n\). Then we vote for the best thought as our strategy \(s\). This strategy \(s\) along with \(q\) are inputted into \(q_n\) to produce an enhanced query \(p\). Based on \(p\), information \(Cr\) is retrieved from the context \(C\), and both \(q\) and \(Cr\) are fed into an LLM (MLP) to generate a response \(r\). A feedback system then evaluates \(r\), yielding an evaluation \(e\). This evaluation \(e\) is subsequently fed to \(q_m\) for updated strategy in the next iteration.

\paragraph{GNN-Att-MLP.} The GNN-Att-MLP setup is similar to the Tree-Att-MLP configuration but introduces the capability to develop deeper graph-based strategies based on previous ones. Initially, for a given query \(q\), we generate a series of thoughts \(T_1, \ldots, T_n\) and identify the best one among them, denoted as \(T_b\). Subsequently, we construct a second layer of thoughts \(T_1', \ldots, T_n'\) based on the initial set \(T_1, \ldots, T_n\). The most effective strategy \(s\) is then selected from among \(T_b\) and \(T_1', \ldots, T_n'\). This chosen strategy \(s\) undergoes the same procedure as described in the Tree-Att-MLP setting.

\section{Automatic Optimization } \label{opt}

\begin{table}[ht]
\vspace{-3mm}
\caption{Optimization range of each hyperparameter.}
\label{optimization range}
\centering
\begin{tabular}{p{0.3\columnwidth}p{0.25\columnwidth}>{\centering\arraybackslash}p{0.15\columnwidth}p{0.10\columnwidth}}
\Xhline{1pt}
& \textbf{Range}  \\
\midrule
Output Size $m_1$ & [64, 128, 192, 256, 320]\\
\midrule
Output Size $m_2$ & [64, 128, 192, 256, 320] \\
\midrule
Output Size $m_3$ &[64, 128, 192, 256, 320] \\
\midrule
Head Num $h$  & [1, 2, 3, 4, 5] \\
\midrule
Retrieve Num $n$  & [1, 2, 3, 4, 5] \\
\Xhline{1pt}
\end{tabular}
\vspace{-4mm}
\end{table}

\begin{figure}[ht]
    \centering
    \includegraphics[width=\linewidth]{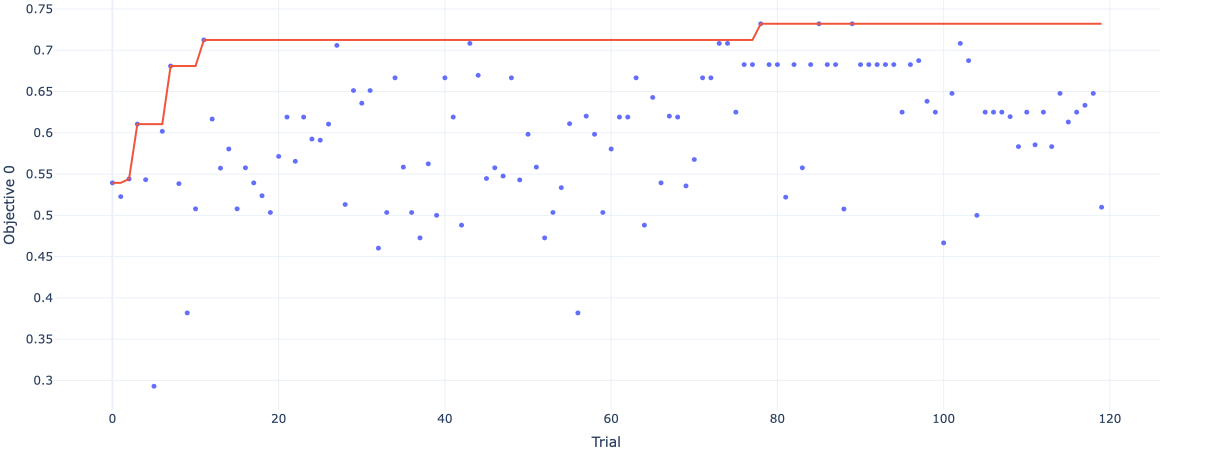}
    \caption{\textbf{The optimization process on  transformer}. The x-axis represents the number of trials. For each trial, we will get a set of $pr, F1$; the y-axis represents the $F1$ value. 
}
\label{fig:NAS}
\end{figure}

\begin{table}[t]
\caption{\textbf{Overview of Datasets.}}
\label{tab:dataset}
\centering
\renewcommand{\arraystretch}{1.2}
\setlength{\tabcolsep}{6pt}
\begin{tabular}{lll}
\toprule
\textbf{Dataset} & \textbf{Task Type} & \textbf{Cases} \\
\midrule
\multicolumn{3}{c}{\textbf{Without External Knowledge}} \\
\midrule
TSP           & Optimization          & 1500    \\
GSM8K         & Math Problem          & 1{,}319 \\
\midrule
\multicolumn{3}{c}{\textbf{With External Knowledge}} \\
\midrule
StrategyQA    & Commonsense Reasoning & 2{,}290 \\
Related-multi & Summary               & 1000    \\
\bottomrule
\end{tabular}
\end{table}
